\documentclass[10pt,journal,compsoc]{IEEEtran}
\ifCLASSOPTIONcompsoc
  \usepackage[nocompress]{cite}
\else
  \usepackage{cite}
\fi
\ifCLASSINFOpdf
\else
\fi
\usepackage{times}
\usepackage{epsfig}
\usepackage{graphicx}
\usepackage{amsmath}
\usepackage{amssymb}
\usepackage{subfigure}
\usepackage{multirow}
\usepackage{caption}
\begin{document}
%
\title{Analyzing the Affect of a Group of People Using Multi-modal Framework}
%
%
%
%

\author{Xiaohua Huang, Abhinav Dhall, Xin Liu, Guoying Zhao, Jingang Shi, Roland Goecke and Matti Pietik\"{a}inen
\IEEEcompsocitemizethanks{\IEEEcompsocthanksitem X. Huang, X. Liu, G. Zhao, J. Shi and M. Pietik\"{a}inen are with the Center for Machine Vision and Signal Analysis, University of Oulu, Finland.\protect\\
E-mail: xiaohua.huang@ee.oulu.fi, xliu@ee.oulu.fi, gyzhao@ee.oulu.fi, jingang.shi@hotmail.com, mkp@ee.oulu.fi
\IEEEcompsocthanksitem A. Dhall is with the Computational Health Informatics Lab, University of Waterloo, Canada.\protect\\
E-mail: Abhinav.dhall@uwaterloo.ca

\IEEEcompsocthanksitem R. Goecke is with Human-Centred Technology Research Centre, University of Canberra, Australia.\protect\\
E-mail: Roland.Goecke@canberra.edu.au}
\thanks{This work has been submitted to the IEEE for possible publication. Copyright may be transferred without notice, after which this version may no longer be accessible}}

%
%

\markboth{Journal of \LaTeX\ Class Files,~Vol.~14, No.~8, August~2016}%
{Shell \MakeLowercase{\textit{et al.}}: Bare Demo of IEEEtran.cls for Computer Society Journals}
%



\IEEEtitleabstractindextext{%
\begin{abstract}
Millions of images on the web enable us to explore images from social events such as a family party, thus it is of interest to understand and model the affect exhibited by a group of people in images. But analysis of the affect expressed by multiple people is challenging due to varied indoor and outdoor settings, and interactions taking place between various numbers of people. A few existing works on Group-level Emotion Recognition (GER) have investigated on face-level information. Due to the challenging environments, face may not provide enough information to GER. Relatively few studies have investigated multi-modal GER. Therefore, we propose a novel multi-modal approach based on a new feature description for understanding emotional state of a group of people in an image. In this paper, we firstly exploit three kinds of rich information containing face, upperbody and scene in a group-level image. Furthermore, in order to integrate multiple person's information in a group-level image, we propose an information aggregation method to generate three features for face, upperbody and scene, respectively. We fuse face, upperbody and scene information for robustness of GER against the challenging environments. Intensive experiments are performed on two challenging group-level emotion databases to investigate the role of face, upperbody and scene as well as multi-modal framework. Experimental results demonstrate that our framework achieves very promising performance for GER.
\end{abstract}

}

\maketitle

\IEEEdisplaynontitleabstractindextext

%
\IEEEpeerreviewmaketitle

\IEEEraisesectionheading{\section{Introduction}\label{sec:introduction}}

Automatic emotion recognition `in the wild' has received a lot of attention in the recent years. `in the wild' here applies to the attributes such as different backgrounds, head motion, illumination conditions, occlusion and presence of multiple people. This paper deals with the scenario, when a group of people are present in an image. It considers the example of a social event like a convocation, where a group of people get together. In scenarios like this a large number of images are captured. In order to infer the perceived mood\footnote{Mood and emotion are used interchangeably in this paper.} of the group of people in an image, we propose a multi-modal approach based on face-level, body-level and scene-level cues.

\begin{figure}[t]
	\centering
	\includegraphics[width=\linewidth]{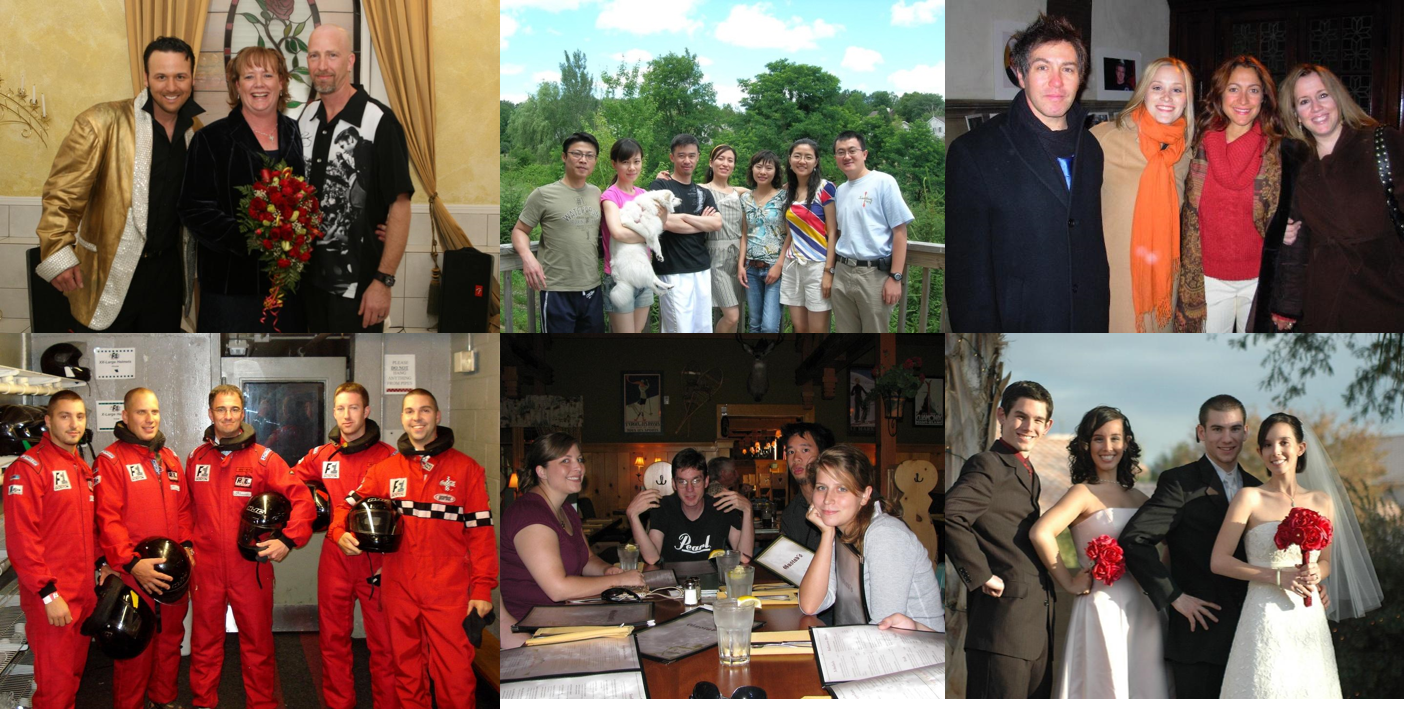}
	\captionsetup{justification=centering}
	\caption{Images of a group of people in social events from HAPPEI database~\cite{Dhall2015}.}
	\label{fig:example}	
\end{figure}

The presence of large pool of data on the Internet, enables us to explore the images containing multiple people (for example Figure~\ref{fig:example}). However, so far relatively little research has examined automatic group emotion in an image. To advance the research in affective computing, it is indeed of interest to understand and model the affect exhibited by a group of people in images. The analysis of group emotion of people in images has various applications such as image management and retrieval - photo album creation\footnote{http://and-fujifilm.jp/en/album/}~\cite{Dhall2010}, key-frame detection and event detection~\cite{Vandal2015}. 

Social psychology studies suggest that group emotion can be conceptualized in different ways. Generally, group emotion can be represented by pairing the bottom-up and top-down approaches~\cite{Barsade1998,Kelly2001}. In the bottom-up methods, group emotion information is inferred by using the subject's attributes. In~\cite{Hernandez2012}, a bottom-up approach, in which it averages over the smiles of multiple people, was proposed to infer the mood of the passerby. The top-down techniques consider external factors to the group and members such as the effect of the scene. In~\cite{Gallagher2009}, Gallagher and Chen proposed contextual features based on the group structure for computing the age and gender of individuals. However, for group affective analysis, a sole approach, such as the bottom-up or top-down, may lose some interesting information. For example, Gallagher and Chen~\cite{Gallagher2009} didn't consider the individual's attribute. In one of the earliest works in automatic group-level affect analysis, Dhall et al.~\cite{dhall2013finding} proposed a hybrid model, which takes into consideration the top-down and bottom-up components of a group. Top-down here means the effect of the scene and group structure on the group and further of the perception of the affect of the group. The scene is a representation of the event and surroundings of a group. An example of the effect of scene can be a scenario where the same joke being told to a group of people sitting in a cafe and to a group of people in a official meeting room may result into a different response (affect). The bottom-up component here is the group members and their attributes such as spontaneous expressions, clothes, age and gender etc. Further they~\cite{dhall2013finding} explore a topic model based approach, which takes into consideration the contribution of the members of the group and the effect of the neighbours. The top-down and bottom-up attributes are learned from a survey, where subjects were asked to rate the perceived affect of a group of people in an image \cite{Dhall2015}. The models here are inspired by the works in social psychology, where group emotions are conceptualized as pairing of the bottom-up and top-down approaches~\cite{Barsade1998,Kelly2001}. Group Expression Models (\textbf{GEM})~\cite{Dhall2015,Huang2015} considered both face-level expressions and the neighbor's effect. Both of them referred global attributes such as the effect of neighboring group members as the top-down component, and local attributes such as an individual's feature as the bottom-up component.

The nature of the images present on the Internet pose challenges due to the presence of illumination variations, head and body pose changes, occlusions, naturalistic and varied indoor and outdoor settings, and interactions taking place between various numbers of people. Existing methodologies on Group-level Emotion Recognition (\textbf{GER})~\cite{Dhall2015,Huang2015} suffer from the failure of face processing, when the face processing pipeline is noisy due to the presence of the challenges discussed above. As we learn from the survey \cite{Dhall2015}, face only may not be the only attribute for understanding the group-level affect. Recent researches demonstrate multiple sources increase the robustness of the system~\cite{AffectiveState, Multimodal, Multimodal2}. Therefore, in this paper we use multiple sources for our system. Specifically, we incorporate the body expression and scene information in the proposed pipeline. From a computer vision perspective it is intuitive to take the scene-level descriptor and body expression features into consideration. Recently, two approaches have been proposed to combine multiple sources for GER~\cite{Dhall2015b,Mou2015}. Dhall et al.~\cite{Dhall2015b} exploited scene information as an additional source for GER. Specifically, they used bag-of-word method to encode facial action unit and face features in a group-level image. Then, they employed two scene descriptors to extract scene features. The combination of face and scene obtained the promising ability. Recently, researchers have reported that body expressions are of significant importance when displaying and understanding expressions~\cite{Joshi2013,Hoai2014,Body}. Motivated by~\cite{Joshi2013,Hoai2014,Body}, another interesting multi-modal work~\cite{Mou2015} combined face and body information to predict the valence and arousal of a group of people. Both works demonstrate that multi-modal framework can perform better than a uni-modal approach. However, both works~\cite{Dhall2015b,Mou2015} have their limitations. For example, in~\cite{Dhall2015b} they did not consider body information in their model; Mou et al.~\cite{Mou2015} ignored scene information and experimented on specific groups based on the fixed number of faces and bodies. Therefore, we propose a new multi-modal approach containing face, body and scene for GER.

Feature extraction from an image containing multiple people is still an open problem for GER. In other words, we aim to use one feature vector for representing multiple persons' information in a group-level image. In the GEMs for GER~\cite{Dhall2015,Huang2015}, they firstly extracted features from the detected faces of group-level images. Then, based on these facial features they constructed group expression model for group-level images. However, they did not really extract the feature from a group-level image. Although Dhall et al.~\cite{Dhall2015b} used bag-of-word to accumulate a histogram from multiple faces for a group-level image, the feature obtained is very sparse and not stable to mis-alignment caused by e.g., head pose change. Recently, an interesting Fisher Vector (\textbf{FV}) method is used describe the face feature by encoding scale-invariant feature transform feature of pixels of an image~\cite{Simonyan2013}. FV is also used as image representation for image retrieval by constructing a bag-of windows from a set of few hundred windows~\cite{Uricchio2015}. In our paper, motivated by~\cite{Simonyan2013,Uricchio2015}, we propose information aggregation to make the features more compact for GER in the wild. On the other hand, our proposed information aggregation method can generate three feature vectors for face, body and scene in a group-level image, leading to the use of some classical feature fusion approaches.

The \textbf{key-contributions} of this paper are as follows: (1) Three modalities- face, upperbody and scene are explored for the task of group-level emotion recognition; (2) A super-pixels based approach is explored for analysing the scene; (3) A robust multi-modal approach is exploited to infer emotion state of a group of people in an image; (4) Extensive experiments are performed on two `in the wild' databases.

To explain the concepts in our approach, the paper is organized as follows: In Section~\ref{sec:database}, we describe two `in the wild' databases used in the paper. In Section~\ref{sec:extraction}, we discuss face, upperbody and scene features descriptor extraction. In Section~\ref{sec:FV}, we encode face, upperbody and scene information for representing group-level image. In Section~\ref{sec:multimodal}, we present our multi-modal framework. In Section~\ref{sec:experiment}, we present the results of face, upperbody and scene as well as the multi-modal approach. Finally, we draw our conclusions in Section~\ref{sec:conclusion}.

\section{Database description}
\label{sec:database}
As far as we know, there are only a few group-level emotion recognition databases. Two of them are available to the research. In this paper, we focus on two `in the wild' databases including The HAPpy PEople Images (\textbf{HAPPEI}) and Group AFFect (\textbf{GAFF}) databases. 

\textbf{HAPPEI database}: It was collected from Flickr by Dhall et al.~\cite{Dhall2015}, in which it contains 2,886 images. All images were annotated with a group level mood intensity. Moreover, in the 2,886 images, 8,500 faces were manually annotated for face level happiness intensity, occlusion intensity and pose by four human labellers, who annotated different images. The mood was represented by the happiness intensity corresponding to six stages of happiness (0-5), i.e., Neutral, Small smile, Large smile, Small laugh, Large laugh and Thrilled. In this database, the labels are based on the perception of the labellers. The aim of this database in~\cite{Dhall2015} is to infer the perceived group mood as closely as possible to human observers. An interesting work of this database is to estimate the happiness intensity of group-level images.

\textbf{GAFF database}: Dhall et al.~\cite{Dhall2015b} extend the HAPPEI database from positive affect only~\cite{Dhall2015} to a wider amount of emotion (Positive, Neutral and Negative) of group of people. They firstly developed a user study in order to understand attributes which affect the perception of affective state of a group. Then they acquired GAFF database~\cite{Dhall2015b} by first searching Flickr and Google images for images related to keywords, which describe groups and events. All images are labelled as three emotion categories (Positive, Neutral and Negative). GAFF database aims to classify group-level image into emotion category.

\section{Information extraction}
\label{sec:extraction}

Analysis of the affect expressed by multiple people is challenging due to the tough situations such as head and body pose variations. Recently, multi-modal emotion recognition for one person is gaining ground~\cite{AffectiveState,Multimodal,Multimodal2}. A group-level image may contain face, upperbody and scene (See Figures~\ref{fig:example}~\ref{fig:HAPPEI}~\ref{fig:GAFF}), which contribute towards the perception of emotion of the group. Thus, we explore face, upperbody and scene for group-level emotion recognition. 

For easy understanding, we define the detected object such as face as sub-image. Considering local information, we segment a sub-image into numerous local regions. For each local region, we use a feature descriptor to extract its information. In this section, we will introduce information extraction for a local region from face, upperbody and scene, respectively.

\subsection{Facial features}

The Riesz transform~\cite{Felsberg2001} has attracted much interest from researchers in the field of face analysis~\cite{Zhang2012}. According to the intrinsic dimension theorem~\cite{Zetzsche1990}, the 1st-order Riesz transform is designed for an intrinsic 1D signal, and the higher-order Riesz transforms is used to characterize the intrinsic 2D local structures. For extracting facial features, we employ Riesz-based Volume Local Binary Pattern (\textbf{RVLBP})~\cite{Huang2015}. A facial image is divided into $m\times n$ overlapped blocks. Then, RVLBP is used for each block.

\begin{figure*}[t]
	\centering
	\includegraphics[width=\linewidth]{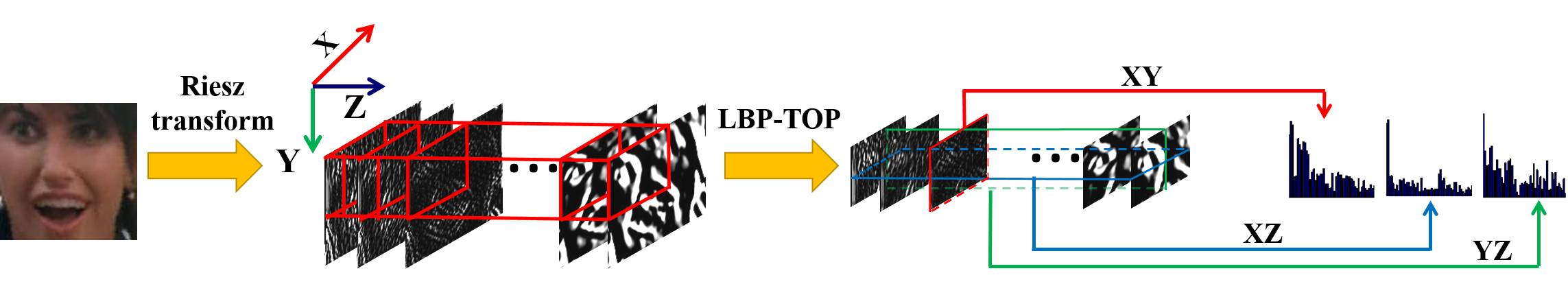}
	\caption{Feature extraction based on $R_x$ Reisz component on one block.}
	\captionsetup{justification=centering}
	\label{fig:RVLBP}	
\end{figure*}

Following~\cite{Huang2015}, we choose the commonly utilized log-Gabor filter~\cite{Fischer2007} with 5 scale and 8 orientations. With the log-Gabor filter $g$, in the case of a 2D image, $\mathbf{X}=(x,y)$, the 1st-order Riesz transforms are expressed as:
\begin{equation}
\label{eqn:hx}
h_x(\mathbf{X})=g*\frac{x}{2\pi\arrowvert \mathbf{X}\arrowvert^3},~~~~\\
h_y(\mathbf{X})=g*\frac{y}{2\pi\arrowvert \mathbf{X}\arrowvert^3},
\end{equation}	
and the 2nd-order ones are represented by:	
\begin{eqnarray}
\label{eqn:hxx}
h_{xx}(\mathbf{X})&=& h_x\{h_x\}(\mathbf{X})=h_x(\mathbf{X})*h_x(\mathbf{X}),\\
\label{eqn:hyy}
h_{yy}(\mathbf{X})&=& h_y\{h_y\}(\mathbf{X})=h_y(\mathbf{X})*h_y(\mathbf{X}), \\
\label{eqn:hxy}
h_{xy}(\mathbf{X})&=&h_x\{h_y\}(\mathbf{X})=h_x(\mathbf{X})*h_y(\mathbf{X}).
\end{eqnarray}

For the $p$-th region $I(x,y)$, we will obtain new Riesz faces for $h_m(\mathbf{X})$ as:
\begin{equation}
\label{eqn:VolumeRiesz}
R_{t}=I(x,y)*[h_{t}(\mathbf{X})_{1}^{1},~\ldots,h_{t}(\mathbf{X})_{5}^{8}], 
\end{equation}
where $t$ is one of $x$, $y$, $xx$, $xy$ and $yy$.

We view volume-based Riesz faces $R_{t}$ as a video sequence. Motivated by~\cite{FacialActions,Zhao2007}, we subsequently employ the Local Binary Pattern from Three Orthogonal Planes operator (\textbf{LBP-TOP}) on $R_{t}$. Finally, we combine the results of these planes to represents faces. For each plane, its histogram is computed as
\begin{equation}
H_{j,t}(l)=\sum_{x,y}L(f_{j,t}(x,y)=l),l=0,1,\ldots,Q_{j,t}-1,
\end{equation}
where $L(x)$ is a logical function with $L(x)=1$ if $x$ is true and $L(x)=0$ otherwise; $f_{j,t}(.)$ expresses the Local Binary Pattern (\textbf{LBP}) codes in the $j$-th plane ($j=0: XY; 1: XZ; 2: YZ$), and $Q_j$ is the bin number of the LBP code in the $j$-th plane. Finally, the histograms $H_{XY,t}$, $H_{XZ,t}$ and $H_{YZ,t}$ are concatenated into one feature vector $H_t$. The procedure is shown in Figure~\ref{fig:RVLBP}. 

Since we have five components for Riesz faces, we employ the above-mentioned procedure for each one. For the $p$-th block, its feature $\Omega_p$ is represented by concatenating $H_x, H_y, H_{xx}, H_{xy}, H_{yy}$.

\begin{figure}[t]
	\centering
	\includegraphics[width=\linewidth]{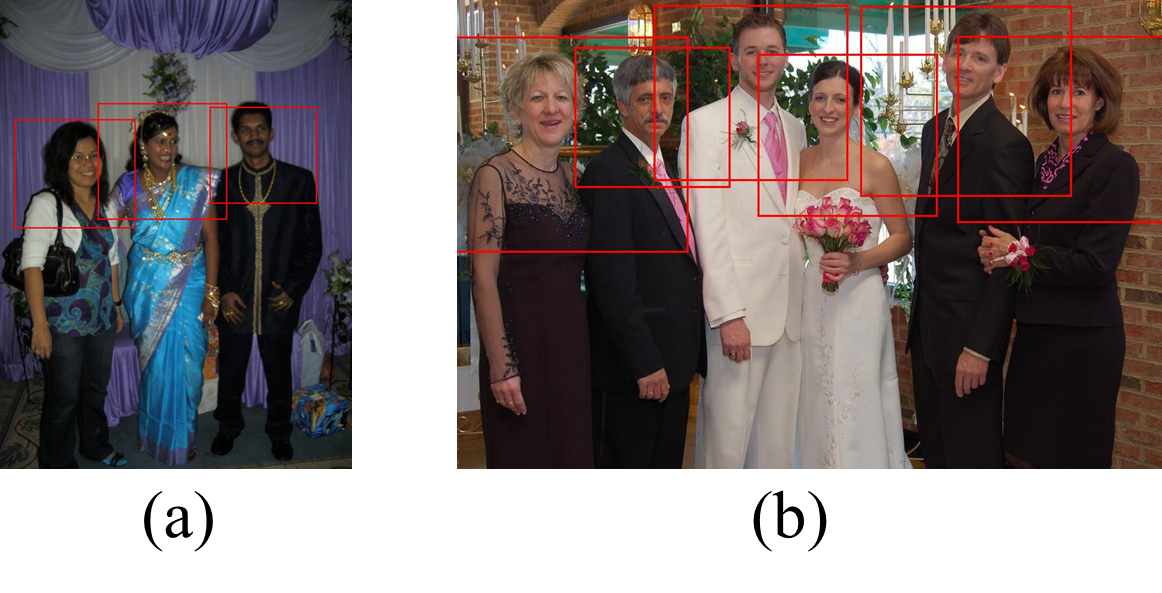}
	\captionsetup{justification=centering}
	\vspace{-20pt}
	\caption{Upperbody detection results on two images of a group of people.}
	\label{fig:upperbody}	
\end{figure}

\subsection{Upperbody features}

Due to low resolution, size and variations in pose and lighting, in naturalistic settings face might provide less reliable evidence for analysing affect. Recently, researchers have reported that body expression are of significant importance when displaying and observing expressions and emotions~\cite{Joshi2013,Hoai2014,Mou2015, Body}. Joshi et al.~\cite{Joshi2013} used the key points of full body for depression analysis. Mou et al.~\cite{Mou2015} implemented the body information for group-level emotion recognition. Therefore, we extract body features as additional source of information for group-level emotion recognition. 

For each person in the image, an upper-body rectangle is detected by using face detection~\cite{Zhu2012} and upperbody detection~\cite{Eichner2009} (see Figure~\ref{fig:upperbody}). It contains face and shoulder information, which may provide interesting information to group-level emotion recognition. However, as seen in Figure~\ref{fig:upperbody}, it is found that the presence of varied backgrounds, illumination change and partial occlusion in challenging conditions may make the group-level emotion recognition even more difficult. So we need to perform feature augmentation by computing low-level features on the upperbody region.

As a dense version of dominating Scale-Invariant Feature Transform (\textbf{SIFT}) feature, Pyramid of Histogram of Gradient (\textbf{PHOG})~\cite{Bosch2007} has shown great success in human upperbody body estimation~\cite{Weinrich2012} and human detection~\cite{Dalal2005}. HOG has been widely accepted as one of the best features to capture the edge or local shape information. On the other hand, the LBP operator~\cite{Abonen2006} is an exceptional texture descriptors. The LBP is highly discriminative and its key advantages, namely its invariance to monotonic gray-level changes and computational efficiency, make it suitable for image analysis tasks such as pose estimation. The combination of PHOG and LBP descriptors has been demonstrated that they can robustly describe the body information to the challenging background~\cite{Wang2009}. HOG performs poorly when the background is clustered with noisy edges. LBP is complementary in this aspect. It can filter out noises using the concept of uniform pattern. The appearance of human upperbody can be better described if edge/local shape information and the texture information are combined. In our method, we divide the detected upperbody image into $m\times n$ overlapped blocks with overlapped ratio 0.7. For the $p$-th block, PHOG is computed by applying edge detection on a sub-region, followed by a histogram computation in a pyramid fashion. For computing LBP, local binary patterns are computed over each pixel which is neighbouring to eight pixels. The feature combination of these two features $\Omega_p$ for the $p$-th block is robust to scale and illumination changes.

\subsection{Scene-level information}
\label{sec:scene}
Group-level image contains much interesting information such as face and cloth. As we know, scene information has been investigated in scene image classification~\cite{Oliva2001, Lazebnik2006}. However, there is few research to use scene-level information for emotion recognition. It is found that background may provide complementary information to group-level emotion recognition. According to the survey in~\cite{Dhall2015b}, attributes such as pleasant scene (background/situation) and cloth affect the perception of human beings to the affect of a group of people in an image. Therefore, we exploit the usefulness of scene analysis features as the global information to our multi-modal framework. 

As we know, two widely used scene analysis descriptors including Census transform histogram (\textbf{CENTRIST})~\cite{Wu2011} and GIST~\cite{Oliva2001} have been employed by~\cite{Dhall2015b} to analyse the emotion of a group of people. However, CENTRIST and GIST used in~\cite{Dhall2015b} model the scene at holistic level. Therefore, we aim to consider the scene at local level. The scene includes many objects and complicated background. It is not reasonable to divide scene image into several blocks, which is the same way to face and upperbody. In this way it will destroy semantic information of scene. Instead, superpixels~\cite{Li2015} enable us to explore the scene feature on a semantically meaningful subregion. 

\begin{figure*}[t!]
	\centering
	\subfigure[]{
		\label{fig:original}
		\includegraphics[width=0.3\linewidth]{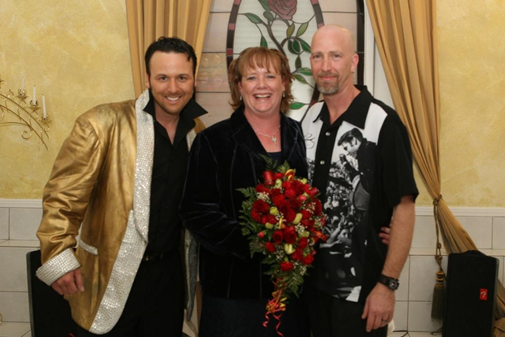}}
	\subfigure[]{
		\label{fig:boundary}
		\includegraphics[width=0.3\linewidth]{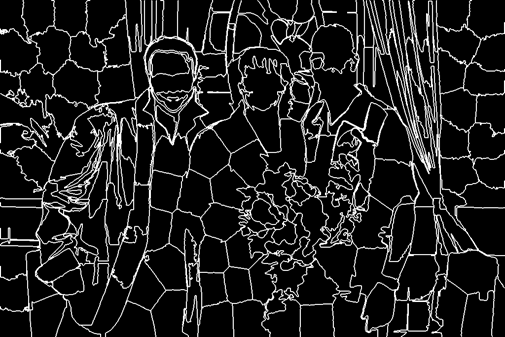}}
	\subfigure[]{
		\label{fig:segmentation}
		\includegraphics[width=0.3\linewidth]{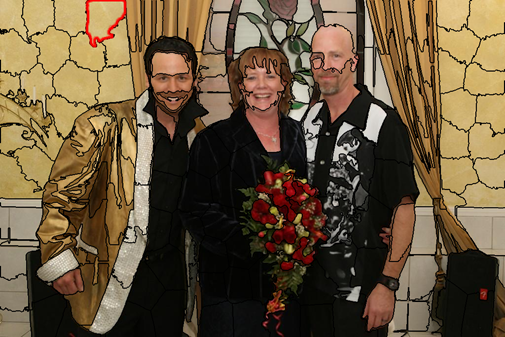}}
		\subfigure[]{
	\label{fig:scene}
		\includegraphics[width=0.7\linewidth]{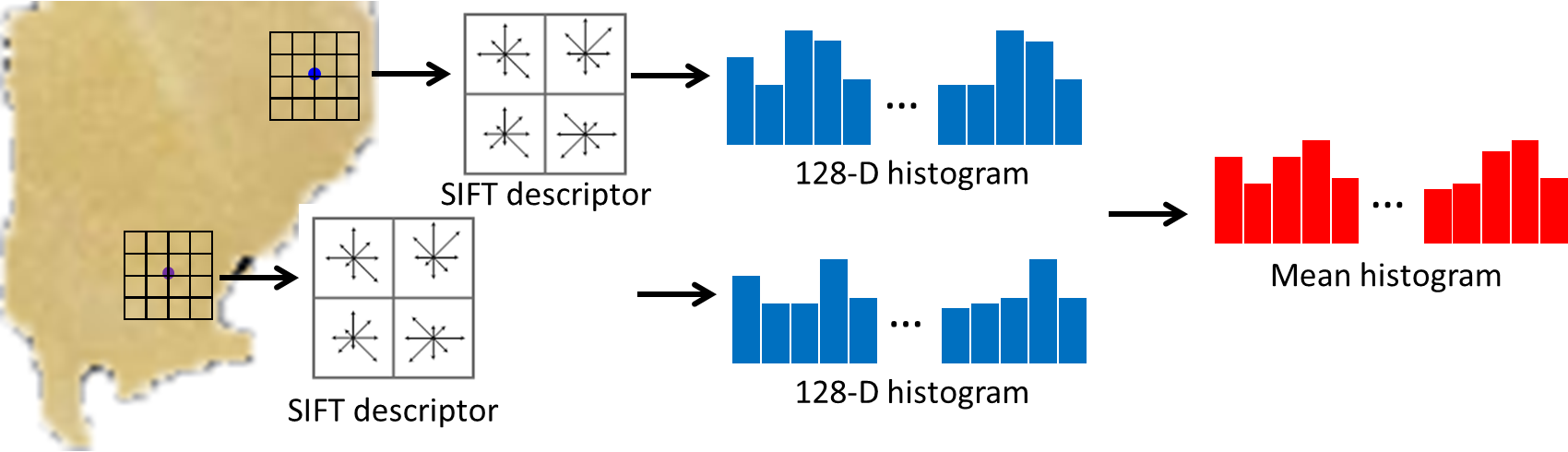}}
	\captionsetup{justification=centering}	
	\caption{Superpixel segmentation result on one group-level image: (a) original image, (b) boundary, (c) region segmentation, in which the red boundary means a superpixel, and (4) an example of feature extraction on a superpixel.}
	\label{fig:superpixel}
\end{figure*}

We segment each group-level image into $N_t$ superpixels by using the Linear Spectral Clustering Superpixels segmentation algorithm in~\cite{Li2015}. We choose an appropriate number $N$ ($N\geq N_t$) so that one superpixel is roughly uniform in color and naturally preserves the boundaries of objects (see Figures~\ref{fig:boundary} and~\ref{fig:segmentation}).

In order to encode appearance information into superpixels, we describe superpixel $S$ by using scale-invariant feature transform (\textbf{SIFT}) descriptor. SIFT has been widely used a local descriptor to characterize local gradient information~\cite{Lowe1999} in object recognition~\cite{Chiu2013}. SIFT has been accepted as one of the best features resistant to common image deformation. In~\cite{Lowe1999}, SIFT descriptor is a sparse feature representation that consists of both feature extraction and detection. In this paper, however, we only use the feature extraction component. Given the $p$-th superpixel $S_p$, for every pixel, we divide its neighborhood ($16\times16$) into a $4\time 4$ cell array, quantize the orientation into 8 bins in each cell, and obtain a 128-dimensional vector as the SIFT representation for a pixel. The procedure is shown in Figure~\ref{fig:scene}. For the $p$-th superpixel $S_p$, its feature can be computed as the average of these per-pixel SIFT descriptors as follows:
\begin{equation}
\Omega_p=\frac{1}{M_p}\sum_{j=1}^{M_p}F_j,
\end{equation}
where $M_p$ is the number of pixel in superpixel $S_p$ and $F_j$ represents SIFT feature of the $j$-th pixel. 

\section{Information aggregation}
\label{sec:FV}

\subsection{Problem formulation}

As previously presented in Section~\ref{sec:extraction}, for the $p$-th region, we obtain features $\Omega^{face}_p$, $\Omega^{body}_p$, $\Omega^{scene}_p$ for face, upperbody and scene, respectively. Concatenating the features of all regions, face, upperbody and scene are represented by $\hat{\Omega}^{face}$, $\hat{\Omega}^{body}$ and $\hat{\Omega}^{scene}$, respectively. We suppose for a group-level image there are $N^{face}$, $N^{body}$ and $N^{scene}$ for the detected faces, upperbodies and superpixels of scene, respectively. Here, we discuss the problem formulation for face. It is the same to upperbody and scene. 

For group-level emotion recognition, how to aggregate $\hat{\Omega}^{face}_1,\ldots,\hat{\Omega}^{face}_{N^{face}}$ into one feature vector $X$ becomes one problem that needs to be resolved. Thus, the problem is defined as follows:
\begin{equation}
	\label{eqn:problem}
	X=\ell(\hat{\Omega}^{face}_1,\ldots,\hat{\Omega}^{face}_{N^{face}}),
\end{equation}
where $\ell$ means the aggregation function.

\subsection{Methodology}

\begin{figure*}[t!]
	\centering
	\subfigure[]{
		\label{fig:GMM}
		\includegraphics[width=0.5\linewidth]{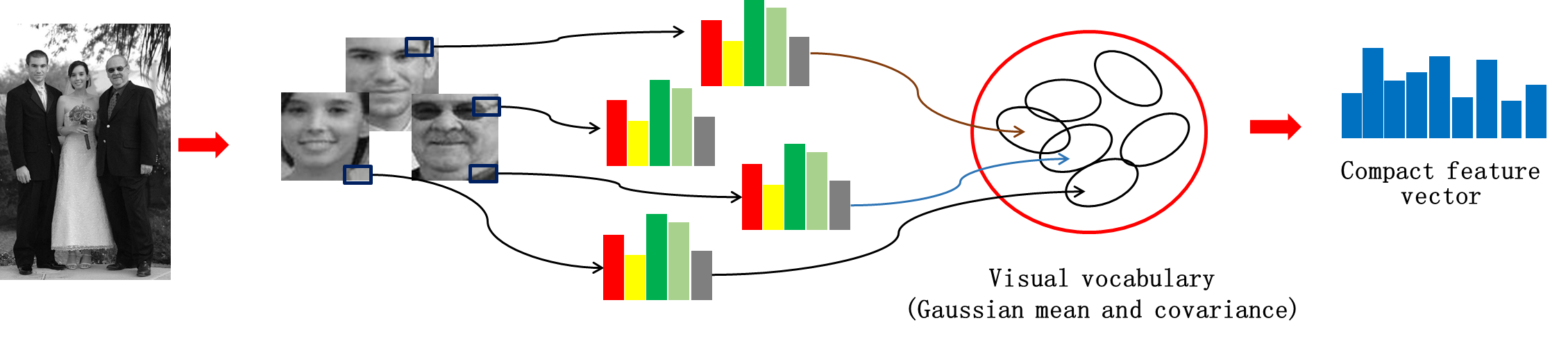}}
	\subfigure[]{
		\label{fig:encoding}
		\includegraphics[width=0.8\linewidth]{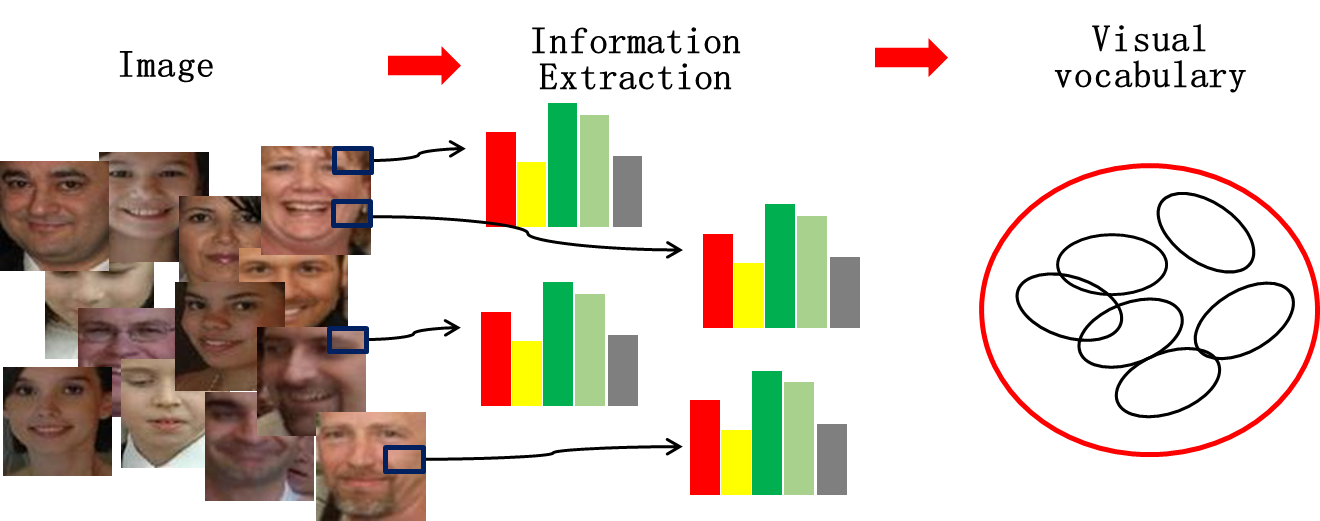}}
	\captionsetup{justification=centering}	
	\caption{Information aggregation procedure for generating a compact feature: (a) visual vocabulary generation and (b) feature encoding.}
	\label{fig:INFA}
\end{figure*}

As we know, for $\ell$, a simple way is to concatenate $\{\hat{\Omega}^{face}_i\}_{i=1}^{N^{face}}$ into one feature vector, but this method requires the number of persons in all group-level images to be the same. An alternative way for $\ell$ is to encode all persons' features by using Bag-of-Word (\textbf{BoW}) method in~\cite{Dhall2015b}, but the feature obtained is very sparse. Both ways seriously limit the application of group-level emotion recognition. Instead, we propose an INFormation Aggregation (\textbf{INFA}) method for $\ell$ as shown in Figure~\ref{fig:INFA} to encode region-based features of multiple persons into a compact feature for a group-level image. As mentioned in Section~\ref{sec:extraction}, for face/upperbody, we divide it into $m\times n$ blocks, while for scene, the regions are obtained by using Superpixel segmentation. In the following, we will present how to obtain $\ell$ for a face. The same procedure is applied to upperbody and scene. 

 Considering local information, Equation~\ref{eqn:problem} is modified as:

\begin{equation}
	\label{eqn:Rproblem}
	X=\ell(\Omega^{face}_1,\ldots,\Omega^{face}_{m\times n\times N^{face}}).
\end{equation}



To encode $\{\Omega^{face}_1,\ldots,\Omega^{face}_{m\times n\times N^{face}}\}$, we exploit the implementation of visual vocabulary. Given the training set $\Omega$, we train visual vocabulary by using a GMM with diagonal covariances under $K$ word size, where word size is defined as the number of Gaussian, and consider derivatives with respect to the Gaussian mean and variance. We obtain visual vocabulary as $\Delta_k=\{w_k, \mu_k, \sigma_k\}$, where $k=1,\ldots,K$, $w_k, \mu_k, \sigma_k$ are the mixture weights, means, and diagonal covariances of the GMM, respectively. Before obtaining GMM, we apply Principal Component Analysis (\textbf{PCA}) on the local regional features to de-correlate features over all regions, where the reduced dimension is $D$. 

For a group-level image, assuming $N_d$ faces are detected, there are $m\times n\times N_d$ regions. Based on $\Delta_k$, this leads to the representation, which captures the average first and second order differences between the regional features and each of the GMM centers:
\begin{equation}
	\label{GMM_mean}
	\Phi_{k}^{(1)}=\frac{1}{m\times n\times N_d\sqrt{w_k}}\sum_{p=1}^{m\times n\times N_d}\alpha_p(k)(\frac{\Omega_p-\mu_k}{\sigma_k}),
\end{equation}
\begin{equation}
	\label{GMM_covariance}
	\Phi_{k}^{(2)}=\frac{1}{m\times n\times N_d\sqrt{2w_k}}\sum_{p=1}^{m\times n\times N_d}\alpha_p(k)(\frac{(\Omega_p-\mu_k)^2}{\sigma_k^2}-1),
\end{equation}
where $\alpha_p(k)$ is the soft assignment weight of the $p$-th feature $\Omega_p$ to the $k$-th word size. For group-level image, it feature $X$ is obtained by stacking the differences: $X=[\Phi_{k}^{(1)}, \Phi_{k}^{(2)},\ldots,\Phi_{K}^{(1)},\Phi_{K}^{(2)}]$.

\section{Revisited localized multiple kernel learning for multimodal framework}
\label{sec:multimodal}

Given a training set with $n$ group images, we have face, upperbody and scene features denoted as $X_r$, where $r\in\{1,2,3\}$, which are obtained by using information aggregation (see Section~\ref{sec:FV}).
The labels are represented as $y_{i}$. We aim to combine face, upperbody and scene features to (1) estimate the happiness intensity of group-level image on the HAPPEI database, which is seen as regression problem, and (2) classify group-level image into three emotion categories (\textit{i.e.,} positive, neutral or negative) on the GAFF database.

\subsection{Problem formulation}

Given face, upperbody and scene for group-level image, it is hoped to combine them in a way that increases the discriminatory power of the features. Generally, concatenating all features is a simple way to perform feature fusion. However, it is not guaranteed that the complementary information will be captured. Possibly, increasing feature dimension will burn the efficient computation. Recently, multiple kernel learning (\textbf{MKL}) has been successfully for image classification~\cite{mkl2015}, audio-video emotion recognition~\cite{Sikka2013} and facial action analysis~\cite{Multimodal3}. Recently, G\"{o}nen et al.~\cite{Gonen2008, Gonen2010} developed the localized multiple kernel learning (\textbf{LMKL}) framework for classification and regression problems. Interestingly, the LMKL algorithm utilized a gating model to select the appropriate kernel function locally and coupled the localizing gating model and the kernel-based classifier in a joint manner. Motivated by LMKL, for fusing face, upperbody and scene for group-level affective state analysis, our problem is formulated by considering the gating function as:
\begin{equation}
\label{eqn:lmkl}
f(x)=\sum_{m=1}^{P}\eta_m\langle \omega_m, F_m(x)\rangle+b
\end{equation}
where $\eta_m$ is the gating function which chooses feature space $m$ as a function of input $x$,  $F_m(x)$ is the mapping function for feature space $m$, $P$ is the number of modalities.

\begin{figure}[t!]
	\centering
	\subfigure[]{
		\label{fig:lmkr}
		\includegraphics[width=\linewidth]{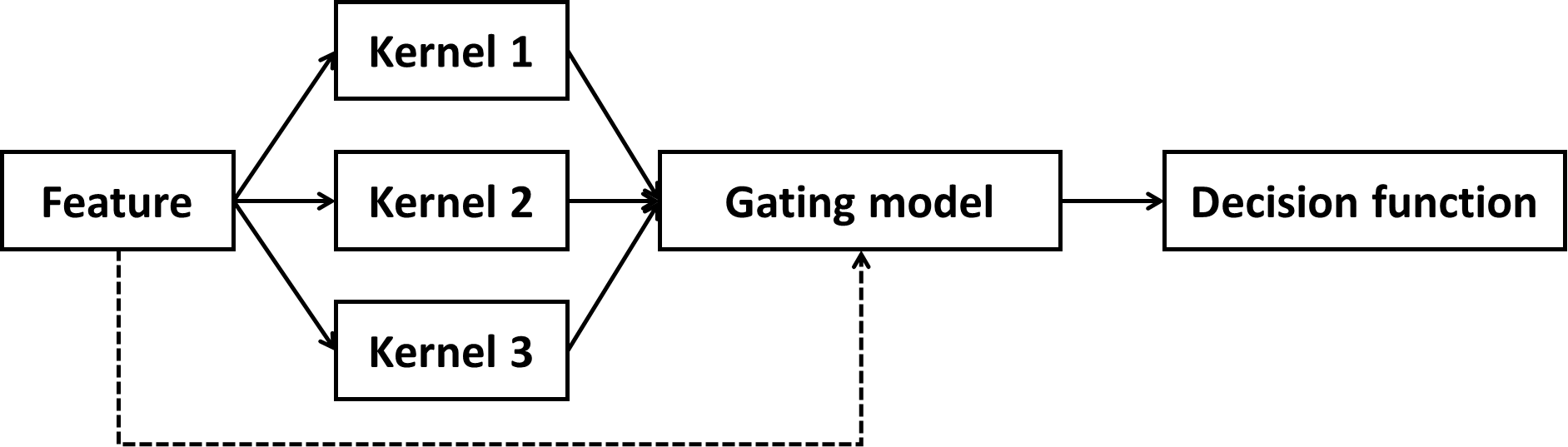}}
	\subfigure[]{
		\label{fig:rlmkr}
		\includegraphics[width=\linewidth]{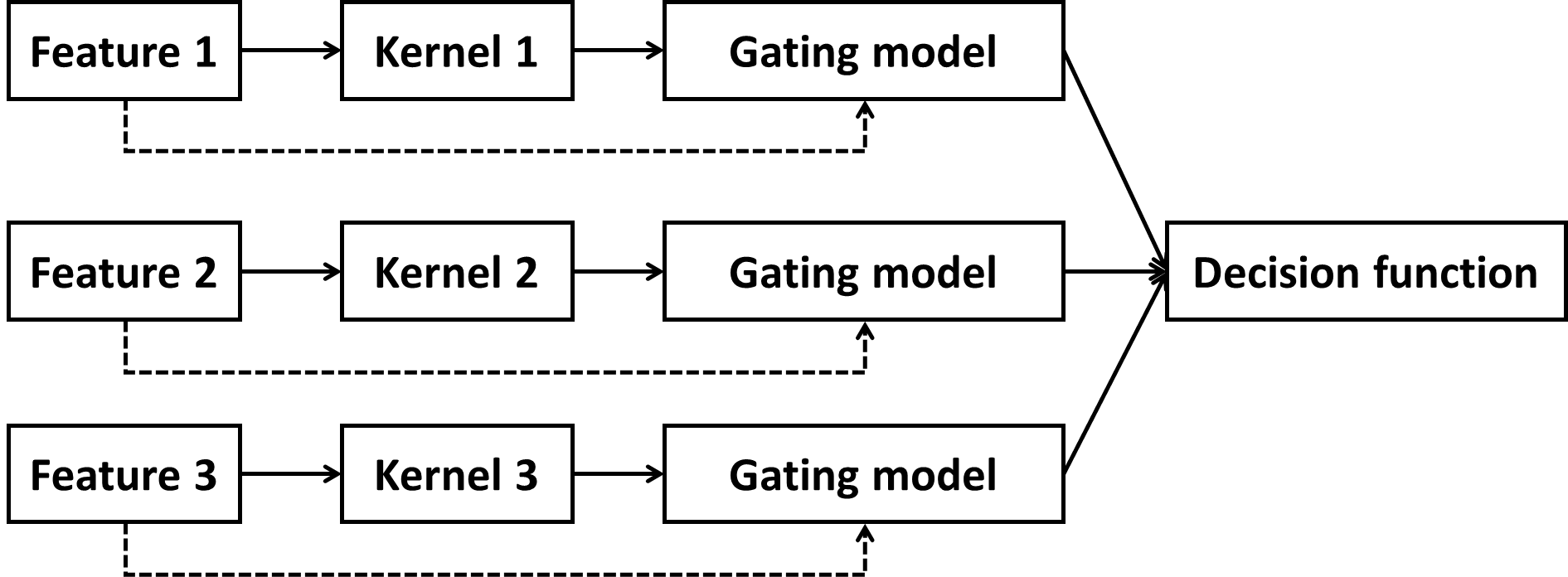}}
	\captionsetup{justification=centering}	
	\caption{The structure framework of (a) LMKR and (b) RLMKR.}
	\label{fig:mkr}
\end{figure}

It is observed that LMKL~\cite{Gonen2008, Gonen2010} investigated the gating model on fusing multiple kernels based on the same input data, as shown in Figure~\ref{fig:lmkr}. The gating model in LMKL is defined up to a set of parameters which are learned from the same input data. Instead, the gating model in feature fusion (in Figure~\ref{fig:rlmkr}) should consider the locality of different modalities. Therefore, we will revisit LMKL for multi-modal framework.

\subsection{Revisited localized multiple kernel learning}

For efficient computation, we firstly introduce the whiten preprocess to each modality. The objective function (Equation~\ref{eqn:lmkl}) on the whitened preprocessed data is rewritten to
\begin{equation}
\label{eqn:rlmkl}
	f(x)=\sum_{r=1}^{P}\eta_r\sum_{i=1}^{n}\alpha_iy_i\text{K}(U_r^Tx_r, U_r^Tx_{r,i})+b,
\end{equation} 
where $\eta_r$ is the $r$-th gating function obtained from $X_r$, $x_{r,i}$ is the $i$-th data sample of the $r$-th feature set $X_r$, $U$ is the whitened matrix by using PCA, $\text{K}$ is the kernel function. For kernel function, we investigate three kinds of kernel function including the linear kernel, the Gaussian kernel and the Histogram intersection (\textbf{HI}) kernel for each modality, which can be defined as following:


(1) Linear kernel  
\begin{equation}
\text{k}(x_i,x_j)=x_ix_j^{T},
\end{equation}
where $T$ represents the transport operator.

(2) Gaussian kernel
\begin{equation}
\text{k}(x_i,x_j)=exp(-\|x_i-x_j\|^2)/s^2,
\end{equation}
where $s$ is the standard deviation.

(3) Histogram intersection kernel:
\begin{equation}
\text{k}(x_i,x_j)=\sum_{h}\text{min}(x_i^{h}, x_j^{h}),
\end{equation}
where $h$ is the index of histogram bins. 

By modifying the original SVM formulation with Equation~\ref{eqn:rlmkl}, we get the following optimization problem for (1) classification:
\begin{equation}
\begin{split}
\label{eqn:svc}
	&\min \frac{1}{2}\sum_{r=1}^{P}\|\omega_r\|^2+C\sum_{i=1}^{n}\xi_i \\
	&\text{w.r.t.}~~ \omega_{r}, b, \xi, \eta_r(X) \\
	&\text{s.t.}~~~ y_i(\sum_{r=1}^{P}\eta_r(X_{r,i})\langle \omega_r, F_r(X_{r,i})\rangle+b)\geq1-\xi_i,\\
	&~~~~~~~~\xi_i\geq0, \forall i, \\
\end{split}
\end{equation}
and for (2) regression:
\begin{equation}
 \begin{split}
 \label{eqn:svr}
 &\min \frac{1}{2}\sum_{r=1}^{P}\|\omega_r\|^2+C\sum_{i=1}^{n}(\xi_i^{+}+\xi_i^{-}) \\
 &\text{w.r.t.}~~ \omega_{r}, b, \xi_i^{+}, \xi_i^{-}, \eta_r(X) \\
 &\text{s.t.}~~~ \epsilon+\xi_i^{+}\geq y_i, \epsilon+\xi_i^{-}\leq y_i, \\
 &~~~~~~~~\xi_i^{+}\geq 0, \xi_i^{-} \geq 0,  \forall i, \\
 \end{split}
 \end{equation}
where $C$ is the regularization parameter, $\{\xi, \xi^{+}, \xi^{-}\}$ are slack variables, and $\epsilon$ is the tube width. 

Following~\cite{Gonen2008}, the optimized problem of classification (Equation~\ref{eqn:svc}) and regression (Equation~\ref{eqn:svr}) is resolved by two-step alternate optimization algorithm. 

(1) For classification problem, the first step is to resolve Equation~\ref{eqn:svc} with respect to $\omega_{r}, b, \xi$ while fixing $\eta_r(X)$, while the second step is to update the parameters of $\eta_r(X)$ using a gradient-descent step calculated from the objective function (Equation~\ref{eqn:svc}). Based on two-step alternate optimization algorithm, the dual formulation is obtained as:

\begin{equation}
\begin{split}
\label{eqn:rlmkl_obj}
	&\max J=\sum_{i=1}^{n}\alpha_i-\frac{1}{2}\alpha_i\alpha_jy_iy_j\text{K}_{\eta}, \\
	&\text{w.r.t.}~~ \alpha \\
	&\text{s.t.}~ \sum_{i=1}^{n}\alpha_iy_i=0~~\text{and}~~C\geq \alpha_i\geq 0, \forall i \\
\end{split}
\end{equation}
where $\text{K}_{\eta}=\sum_{r=1}^{P}\eta_r(X_r)\text{K}_r\eta_r(X_r)$, which is the locally combined kernel matrix, $\text{K}_r=\text{K}(U_r^Tx_r, U_r^Tx_{r,i})$, $\eta_r(X_r)$ is the gating model of $X_r$, 

(2) For regression problem, following the step of classification problem, the dual formulation is easily obtained as:
\begin{equation}
\begin{split}
\label{eqn:rlmkr_obj}
&\max J=\sum_{i=1}^{n}y_i(\alpha_i^{+}-\alpha_i^{-})+\epsilon\sum_{i=1}^{n}y_i(\alpha_i^{+}+\alpha_i^{-}) \\
&~~~~~~~~~-\frac{1}{2}~(\alpha_i^{+}-\alpha_i^{-})(\alpha_j^{+}-\alpha_j^{-})\text{K}_{\eta}, \\
&~\text{w.r.t.}~~ \alpha_i^{+}, \alpha_i^{-} \\
&~\text{s.t.}~ \sum_{i=1}^{n}(\alpha_i^{+}-\alpha_i^{-})=0\\
&~~~~~~~~C\geq \alpha_i^{+}\geq 0, C\geq \alpha_i^{-}\geq 0, \forall i \\
\end{split}
\end{equation}

For the gating model, we implement softmax function which can be expressed as:
\begin{equation}
	\eta_r(X_r)=\frac{exp(\langle v_m,X_r\rangle+v_{r0})}{\sum_{k=1}^{P}exp(\langle v_k,X_k\rangle+v_{k0})},
\end{equation}
where $v_r,v_{r0}$ are the parameters of this gating model and the softmax guarantees nonnegativity.


We can simply use the objective function of Equation~\ref{eqn:rlmkl_obj} or Equation~\ref{eqn:rlmkr_obj} as $J(\eta)$ function is to calculate the gradients of the primal objective with respect to the parameters of $\eta_r(X_r)$. to train the gating model, we take derivatives of $J(\eta)$ with respect to  $v_r,v_{r0}$ and use gradient-descent:
\begin{equation}
\begin{split}
	\frac{\partial J(\eta)}{\partial v_{r0}}&=-\frac{1}{2}\sum_{i=1}^{n}\sum_{j=1}^{n}\sum_{k=1}^{P}\alpha_{i}\alpha_{j}y_iy_j\eta_r\text{K}_r\\
	&\eta_r(\delta_{r,k}-\eta_r+\delta_{r,k}-\eta_r)	
\end{split}
\end{equation}
\begin{equation}
\begin{split}
\frac{\partial J(\eta)}{\partial v_{r}}&=-\frac{1}{2}\sum_{i=1}^{n}\sum_{j=1}^{n}\sum_{k=1}^{P}\alpha_{i}\alpha_{j}y_iy_j\eta_r\text{K}_r\\
	&\eta_r(x_i[\delta_{r,k}-\eta_r]+x_j[\delta_{r,k}-\eta_r])
\end{split}
\end{equation}
where $\delta_{r,k}$ is 1 if $r=k$ and 0 otherwise. After updating the parameters of $\eta_r(X_r)$, we are required to solve a single kernel SVM with $\text{K}_{\eta}$ at each step.

%

\section{Experiments}
\label{sec:experiment}

In previous section, we propose two important works for group-level emotion recognition, including INFormation Aggregation (\textbf{INFA}) to encode information of multiple persons and multi-modal framework. In this section, we will thoroughly evaluate all parameters of INFA and multi-modal framework on the HAPPEI database~\cite{Dhall2015}. Following~\cite{Huang2015}, four-fold cross validation method is used, where 1,500 images are used for training and 500 for testing, repeating 4 times. The task is to predict the intensity level of group-level images, so mean absolute error (\textbf{MAE}) is used as measurement. Finally, we evaluate the generalized ability of our proposed method on GAFF database~\cite{Dhall2015b} using the well-designed parameters of HAPPEI database.

\subsection{Evaluation of INFA to face, upperbody and scene}

Previously mentioned in Section~\ref{sec:FV}, block number of face or upperbody area $m\times n$, PCA dimension $D$ and word size $K$ are three important parameters in INFA. Additionally, different regional features may provide varying performance to INFA. In this experiment, we will firstly focus on the influence of three parameters, various regional features and different encoding methods on face-level information. Subsequently, we will investigate the effect of three parameters on body-level and scene-level information.

\subsubsection{Face}
\label{sec:faceExp}

Firstly, we evaluate the influence of PCA dimension and block number. Three kinds of block number $\{4\times 4$, $8\times 8$, $16\times 16\}$ and six different PCA dimensions $\{16, 32, 64, 128, 256, 512\}$ are considered. We obtain the results under block number and PCA dimension for HAPPEI database as shown in Table~\ref{tab:faceHAPPEI}, where $K=50,100,150$.

\textbf{Block number}: As seen from Table~\ref{tab:faceHAPPEI}, the MAE is favourably decreased using more blocks. It means that GMM has enough training features to learn the feature distribution.  

\textbf{PCA dimension}: In Table~\ref{tab:faceHAPPEI}, increasing PCA dimension improves the performance. The best results are obtained at $D=256$ for various word sizes and block numbers on HAPPEI database. 

\begin{table*}[t!]
	\small
	\centering
		\captionsetup{justification=centering}	
	\caption{Mean absolute error (MAE) of INFA based on face information for various PCA dimension and block number on HAPPEI database, where the bold number is the smallest MAE along the row and the bold number underline is the lowest MAE in the table.}
	\label{tab:faceHAPPEI}
	\begin{center}
		\begin{tabular}{|c|c|c|c|c|c|c|c|}
			\hline
			\multirow{2}{*}{\textbf{Word size}} & \multirow{2}{*}{\textbf{Block Number}} & \multicolumn{6}{c|}{\textbf{PCA Dimension}}\\
			
			\cline{3-8}\multicolumn{1}{|c}{}&\multicolumn{1}{|c}{}&\multicolumn{1}{|c}{16}&\multicolumn{1}{|c}{32}&\multicolumn{1}{|c}{64}&\multicolumn{1}{|c|}{128}&\multicolumn{1}{|c|}{256}&\multicolumn{1}{|c|}{512}\\
			\hline
			50  &  $4\times 4$ & 0.6381 &  0.5873 & 0.5565 & \textbf{0.5488} & 0.5659 & 0.5545\\
			50  & $8\times 8$ & 0.6397 &  0.5951 & 0.5896 & 0.5761 & 0.5556 & \textbf{0.5540}\\
			50  & $16\times 16$ & 0.6094 & 0.5975 & 0.569 & 0.5472 & \textbf{0.5432} & 0.5477 \\ \hline
			100 & $4\times 4$ & 0.6257 & 0.5790 & \textbf{0.5507} & 0.5687 & 0.5584 & 0.5556\\
			100 & $8\times 8$ & 0.6131 & 0.5905 & 0.586 & 0.5508 & \textbf{0.5364} & 0.547 \\
			100 & $16\times 16$ & 0.6052 & 0.5716 & 0.5659 & 0.5452 & \textbf{0.5293} & 0.5318\\ \hline
			150 & $4\times 4$ & 0.6186 & 0.5696 & 0.5681 & 0.5644 & 0.5682 & \textbf{0.5606}\\
			150 & $8\times 8$ & 0.6094 & 0.5962 & 0.5804 & 0.5484 & \textbf{0.5363} & 0.5388 \\
			150 & $16\times 16$ & 0.6014 & 0.5637 & 0.5537 & 0.5381 & \underline{\textbf{0.5255}} & 0.5286\\
			\hline
		\end{tabular}
	\end{center}
\end{table*}

\begin{table}[t!]
	\small
		\captionsetup{justification=centering}	
	\caption{Performance of INFA using different feature descriptors on HAPPEI database.}
	\label{tab:featComp}
	\begin{center}
		\begin{tabular}{|c|c|c|}
			\hline
			\textbf{Algorithm} & Parameters & MAE \\
			\hline
			LBP & $16\times 16$, $D=32$, $W=150$ & 0.6092 \\
			LPQ  & $16\times 16$, $D=128$, $W=160$ & 0.6047 \\
			RVLBP & $16\times 16$, $D=256$, $W=180$ & \textbf{0.5187}\\
			\hline
		\end{tabular}
	\end{center}
\end{table}

\begin{figure}[t!]
	\centering
		\includegraphics[width=\linewidth]{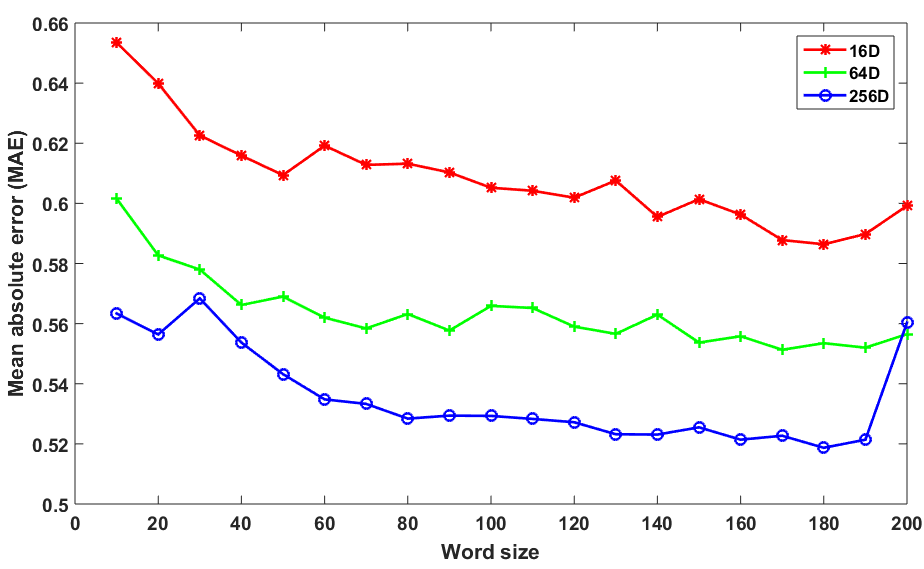}
		\captionsetup{justification=centering}	
	\caption{Performance using INFA based on face information under various word size on HAPPEI database.}
	\label{fig:FaceWordSize}
	
\end{figure}

\textbf{Word size}: Based on the above parameter setup, we furthermore discuss the influence of word size $K$ to INFA. Figure~\ref{fig:FaceWordSize} shows the effect of different $K$ to INFA. As seen from Figure~\ref{fig:FaceWordSize}, the MAE was considerably decreased with the increasing word size for all $D$. INFA method obtains comparative MAE of 0.5187 at $K=180$. We see that the large word size and suitable PCA dimension can provide promising performance to INFA.

We compare RVLBP with LBP and Local Phase Quantization (\textbf{LPQ})~\cite{Ojansivu2008} as region descriptor for INFA. The comparison is shown in Table~\ref{tab:featComp}. Comparing with LBP and LPQ, INFA using RVLBP comparatively outperforms LBP and LPQ. RVLBP provides more useful and discriminative information than LBP and LPQ, since RVLBP incorporates the spatial information and the co-occurrence statistics on frequency and orientation domains of Riesz transform.

Finally, we compare INFA with Bag-of-Word (\textbf{BoW}) and Vector of Locally Aggregated Descriptor (\textbf{VLAD}), where k-means method is used for all regional features for BoW and visual dictionary is obtained by vector quantization for VLAD. Support vector regression (\textbf{SVR}) based on linear kernel is used. Table~\ref{tab:Cluster} shows the MAE of BoW, VLAD and INFA. In Table~\ref{tab:Cluster}, we see that BoW fails to get promising results while VLAD has closely ability to INFA. From this result, INFA can achieve a promising performance compared to two classical clustering methods. 

\begin{table}[t!]
	\small
		\captionsetup{justification=centering}	
	\caption{Comparison of BoW, VLAD and INFA on HAPPEI database.}
	\label{tab:Cluster}
	\begin{center}
		\begin{tabular}{|c|c|c|c|}
			\hline
			 \multirow{2}{*}{\textbf{Measurement}} & \multicolumn{3}{c|}{\textbf{Algorithm}}\\
			
			\cline{2-4}\multicolumn{1}{|c}{} & \multicolumn{1}{|c}{BoW}&\multicolumn{1}{|c}{VLAD} &\multicolumn{1}{|c|}{INFA}\\
			\hline
			 MAE & 0.7015 & 0.6384 & \textbf{0.5187}\\
			\hline
		\end{tabular}
	\end{center}
\end{table}

\subsubsection{Upperbody}

\begin{table*}[t!]
	\small
	\captionsetup{justification=centering}	
	\caption{Mean absolute error of INFA based on upperbody information for various PCA dimension and block number on HAPPEI database, where the bold number is the lowest MAE along the row and the bold number underline is the lowest MAE on all results.}
	\label{tab:BodyHAPPEI}
	\begin{center}
		\begin{tabular}{|c|c|c|c|c|c|c|c|c|}
			\hline
			\multirow{2}{*}{\textbf{Word size}} & \multirow{2}{*}{\textbf{Block Number}} & \multicolumn{7}{c|}{\textbf{PCA Dimension}}\\
			
			\cline{3-9}\multicolumn{1}{|c}{}&\multicolumn{1}{|c}{}&\multicolumn{1}{|c}{16}&\multicolumn{1}{|c}{32}&\multicolumn{1}{|c}{64}&\multicolumn{1}{|c|}{128}&\multicolumn{1}{|c|}{256}&\multicolumn{1}{|c|}{512}&\multicolumn{1}{|c|}{1024}\\
			\hline
			80  &  $4\times 4$ & 0.7576 & 0.7435 & 0.7340 & 0.7355 & \textbf{0.7304} & 0.7616 & 0.7478\\
			80  & $8\times 8$ & 0.7367 & 0.7309 & \textbf{0.7157} & 0.7166 & 0.7227 & 0.7251 & 0.7392\\
			80  & $16\times 16$ & 0.7320 & 0.7235 & \textbf{0.7040} & 0.7174 & 0.7281 & 0.7279 & 0.7383\\ \hline
			120 & $4\times 4$ & 0.7631 & 0.7393 & \textbf{0.7313} & 0.7314 & 0.7340 & 0.7517 & 0.7491\\
			120 & $8\times 8$ & 0.7466 & 0.7241 & \textbf{0.7093} & 0.7131 & 0.7230 & 0.7247 & 0.7171\\
			120 & $16\times 16$ & 0.7210 & 0.7128 & \underline{\textbf{0.6962}} & 0.7029 & 0.7239 & 0.7265 & 0.7282\\ \hline
			160 & $4\times 4$ & 0.7558 & \textbf{0.7245} & 0.7329 & 0.7372 & 0.7385 & 0.7511 & 0.7456\\
			160 & $8\times 8$ & 0.7415 & \textbf{0.7183} & 0.7187 & 0.7209 & 0.7195 & 0.7223 & 0.7424\\
			160 & $16\times 16$ & 0.7104 & 0.7107 & \textbf{0.7082} & 0.7158 & 0.7145 & 0.7251 & 0.7462\\
			\hline
		\end{tabular}
	\end{center}
\end{table*}

The detected upperbody images are resized to $128\times 128$. A three-level PHOG~\cite{Bosch2007} is computed first on the upperbody sub-regions. The number of bins for the histogram representation is set to 10, and orientation range $[0, 360]$. For the LBP descriptor, the number of neighbors is 8 and radius 3 for one sub-region. After concatenation a 1106-dimensional feature vector is obtained for each sub-region. In this section, we aim to evaluate the contribution of body information to group-level emotion recognition. Parameter setup and classification are the same to Section~\ref{sec:faceExp}.

\textbf{PCA dimension and Block number}: Tables~\ref{tab:BodyHAPPEI} shows the results of INFA based on body information with different PCA dimension and block number. As seen from the table, INFA with $16\times 16$ achieves the lowest MAE of 0.6962. It demonstrates that increasing block number can considerably improve the performance of INFA on body information. The results of Tables~\ref{tab:faceHAPPEI} and ~\ref{tab:BodyHAPPEI} consistently demonstrate that more blocks can make GMM more compact to distribution of samples.

\begin{figure}[t!]
	\centering
		\includegraphics[width=\linewidth]{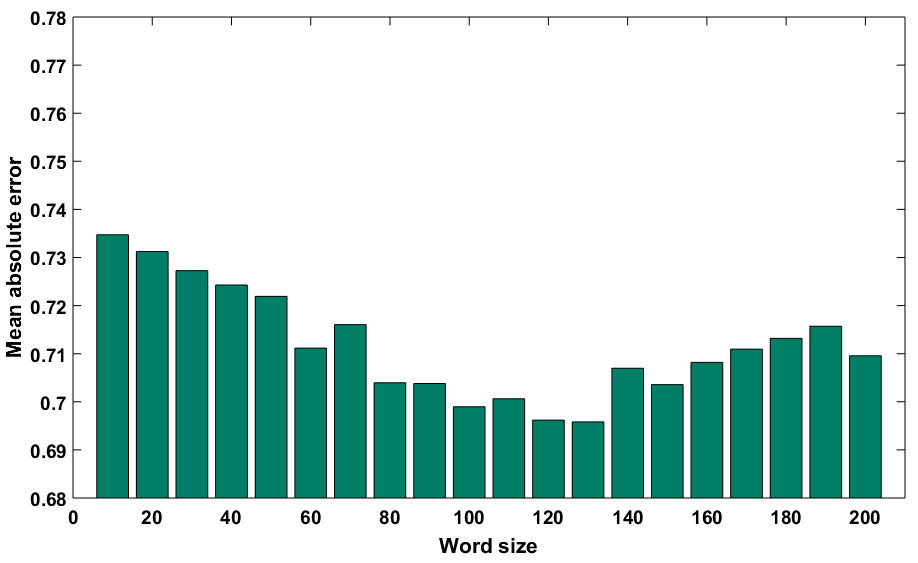}
	\captionsetup{justification=centering}	
	\caption{Performance of INFA based on upperbody information under word size ranging in [10 200] on HAPPEI database.}
	\label{fig:PoseWordSize}
\end{figure}

\textbf{Word size}: We make the experiment to evaluate how varying word size affects INFA based on body information. The effect of $K$ is presented in Figure~\ref{fig:PoseWordSize}. For parameter setup, the block number is set as $16\times 16$ and PCA dimension as 128. With increasing word size, the performance is encouragingly increased, since more word size can make INFA feature more compact. It is observed that INFA obtains 0.6958 of MAE at word size 130 for HAPPEI database. 

Finally, we observe that only using PHOG or LBP achieved the MAE of 0.7571 and 0.7486 at the above-mentioned designed parameters, respectively. It is seen that the combination of PHOG and LBP has complementary information to each other.

\subsubsection{Scene}

\begin{table}[t!]
	\small
	\captionsetup{justification=centering}	
	\caption{Comparison of CENTRIST, GIST and $\text{INFA}_{\text{Scene}}$ on HAPPEI database.}
	\label{tab:sceneComp}
	\begin{center}
		\begin{tabular}{|c|c|c|c|}
			\hline
			\multirow{2}{*}{\textbf{Database}} & \multicolumn{3}{c|}{\textbf{Algorithm}}\\
			
			\cline{2-4}\multicolumn{1}{|c}{}& \multicolumn{1}{|c}{CENTRIST~\cite{Wu2011}}&\multicolumn{1}{|c}{GIST~\cite{Oliva2001}}&\multicolumn{1}{|c|}{$\text{INFA}_{\text{Scene}}$}\\
			\hline
			HAPPEI  & 0.7134 & 0.7371 & 0.7169\\
			\hline
		\end{tabular}
	\end{center}
\end{table}

\begin{figure}[t!]
	\centering

		\includegraphics[width=\linewidth]{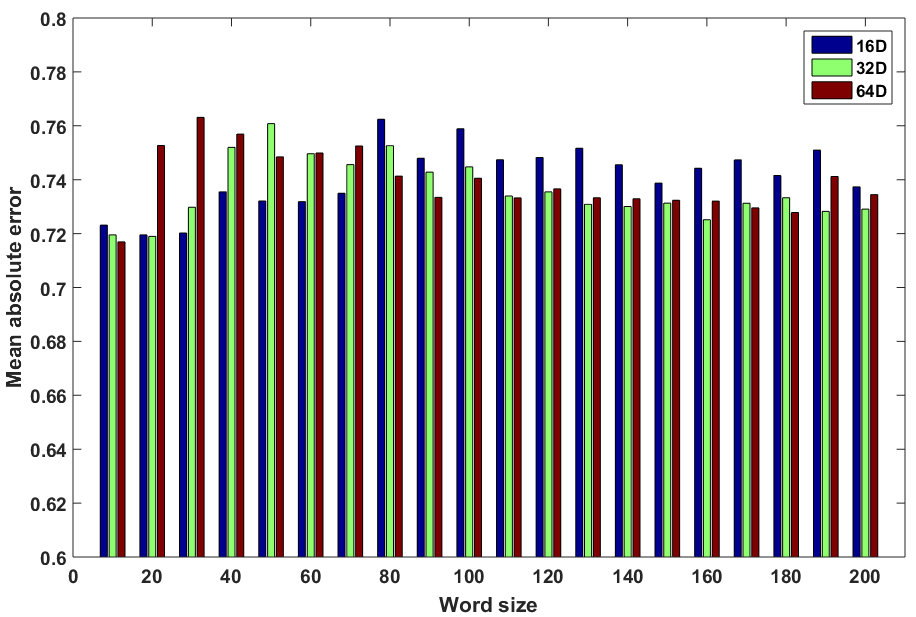}
		\captionsetup{justification=centering}	
	\caption{Performance of INFA using scene information under word size ranging in [10 200] on HAPPEI database.}
	\label{fig:SceneWordSize}
\end{figure}

In this section, we aim to see the benefit of scene for group-level emotion recognition. Previously mentioned in Section~\ref{sec:scene}, block number is eliminated by using superpixel method. Therefore, in this section we discuss how word size and PCA dimension leverage the performance of INFA. The results are shown in Figure~\ref{fig:SceneWordSize}. It is seen that the smallest word size (10) achieves the lowest MAE of 0.7169 for three PCA dimensions.

Following~\cite{Dhall2015b}, we employ CENTRIST and GIST as comparison to our INFA features.  The publicly available GIST implementation\footnote{http://people.csail.mit.edu/torralba/code/spatialenvelope/} is used with its default parameters: Orientations per scale is $[8, 8, 8, 8]$; number of blocks is 4. Similarly for the CENTRIST descriptor, publicly available implementation\footnote{http://github.com/sometimesfood/spact-matlab} is used. The comparative result is shown in Table~\ref{tab:sceneComp}. As seen from this table, CENTRIST works better than INFA but their results are quite close. INFA outperforms GIST.

\subsection{Performance of multi-modal framework}

\begin{table}[t!]
	\small
		\captionsetup{justification=centering}	
	\caption{Designed Parameter and MAE of INFA on HAPPEI database for face, upperbody and scene, where the numbers in the second column represent block number, reduced dimensionality by PCA and word size, respectively}
	\label{tab:INFApara}
	\begin{center}
		\begin{tabular}{|c|c|c|}
			\hline
			Information & Parameter & MAE\\ \hline
			Face & $16\times 16$, 256, 180 & 0.5187 \\ \hline
			Upperbody & $16\times 16$, 128, 130 & 0.6958 \\ \hline
			Scene & -, 64, 10 & 0.7169 \\
			\hline
		\end{tabular}
	\end{center}
\end{table}

According to above-mentioned experimental setup, we design the good parameters with respect to the promising performance for face, upperbody and scene on HAPPEI database. Table~\ref{tab:INFApara} presents the designed parameter in this scenario, where INFA on face, upperbody and scene obtained the mean absolute errors (MAE) of 0.5187, 0.6958 and 0.7169, respectively.

We test the performance of revisited localized multiple kernel learning based on three kernel functions including linear, Gaussian and Histogram Intersection (\textbf{HI}) on Happiness intensity estimation. We choose the optimal parameters of $C$ and $\epsilon$ using 10-fold-cross-validation on training set. For Gaussian kernel, we choose the $s=10$ for the standard deviation. Table~\ref{tab:fusion} shows the results of combining different modalities. It is found that using linear kernel function works better than two other kernel functions. We see that face is improved by adding upperbody information, since upperbody provides the body pose to face. As well, adding scene information can promisingly improve the performance of face, since attributes such as pleasant scene may attract the perception of human beings when they analyse the affect of a group of people in an image. The fusion of face, upperbody and scene information performs the best out of all tested configurations. One based on upperbody and scene performs less than fusion of face and upperbody/scene. It is to understood here, that given the ´in the wild' nature of the images, face detection may fail, so we need to add complementation information, which is obtained from the scene-level and body-level descriptors.

\begin{table}[t!]
	\small
	\captionsetup{justification=centering}	
	\caption{Performance of multi-modal framework based on MKL, where F, P and S represents face, pose and scene, respectively. The bold number underline is the lowest MAE and the bold number is the second lowest MAE in the table.}
	\label{tab:fusion}
	\begin{center}
		\begin{tabular}{|c|c|c|c|c|}
			
			\hline
			\multirow{2}{*}{\textbf{Kernel}} & \multicolumn{4}{c|}{\textbf{Modality}}\\
			
			\cline{2-5}\multicolumn{1}{|c}{}&\multicolumn{1}{|c|}{F+P}&\multicolumn{1}{|c|}{F+S}&\multicolumn{1}{|c|}{P+S} &\multicolumn{1}{c|}{F+P+S}\\
			\hline
			Linear & 0.5010 & 0.4915 & 0.6154 & \underline{\textbf{0.4835}} \\ \hline
			Gaussian & 0.5177 & 0.5242 & 0.6072 & 0.5107 \\ \hline
			HI & 0.5148 & 0.5199 &  0.7273 &  \textbf{0.5002} \\ \hline
			\end{tabular}
	\end{center}
	\vspace{-10pt}
\end{table}

\subsection{Algorithm comparison}

Dhall et al.~\cite{Dhall2015} proposed the Group Expression Model (\textbf{GEM}) based on average, weight and Latent Dirichlet Allocation (\textbf{LDA}) to estimate happiness intensity of a group of people. Huang et al.~\cite{Huang2015} proposed a new GEM based on Continuous Conditional Random Filed (\textbf{CCRF}) for happiness intensity estimation. We compare INFA with all GEM models~\cite{Huang2015}. Table~\ref{tab:algCompHAPPEI} shows MAE of our method and comparisons. As seen from this table, INFA using face information achieves comparative results comparing with all GEM models, increased by 0.0457. But for upperbody and scene, INFA works worse than GEM models based on face. Combining face, upperbody and scene acceptably improves the performance of face since it uses additional information to face from upperbody and scene.

\begin{table}[t!]
	\small
	\captionsetup{justification=centering}	
	\caption{Algorithm comparison on HAPPEI, where results of GEM are from~\cite{Huang2015}. The bold number underline is the lowest MAE and the bold number is the second lowest MAE in the table.}
	\label{tab:algCompHAPPEI}
	\begin{center}
		\begin{tabular}{|c|c|c|c|}
			\hline
			Algorithm & MAE & Algorithm & MAE\\ \hline
			
			$\text{GEM}_{avg}$  & 0.5622 & $\text{INFA}_{Face}$ & \textbf{0.5187} \\ \hline
			$\text{GEM}_{w}$ & 0.5469 & $\text{INFA}_{Body}$ & 0.6958 \\ \hline
			$\text{GEM}_{LDA}$ & 0.5407 & $\text{INFA}_{Scene}$ & 0.7169\\ \hline
			$\text{GEM}_{CCRF}$ & 0.5292 & Multi-modal & \underline{\textbf{0.4835}}\\ \hline	
		\end{tabular}
	\end{center}
	\vspace{-10pt}
\end{table}

From intensive comparisons on HAPPEI database, our multi-modal method achieves a considerable performance for group-level happiness intensity estimation. Additionally, we also show upperbody and scene information contribute their role to analyze emotion state of a group of people.

\subsection{Evaluation of multi-modal system on GAFF database}

Based on the well-designed parameters, we evaluate the performance of our proposed method on GAFF database~\cite{Dhall2015b}. In the experiment, 417 out of 504 images are chosen in our experiments, since face detection failed to work in 87 images. Following~\cite{Dhall2015b}, 213 images are chosen for training set and 204 for test set. Different from HAPPEI database, the task in GAFF database is to classify group-level images into `Positive', `Neutral' or `Negative'. Recognition accuracy is used as a measurement. Using the designed parameters in Table~\ref{tab:INFApara}, INFA obtains the accuracy of 58.33\%, 46.57\% and 42.16\% for face, upperbody and scene information, respectively. We use revisited localized multiple kernel learning to combine face, upperbody and scene. In the implementation, we conduct revisited localized multiple kernel learning based on the one-vs-one classification problem. Our system obtains the accuracy of 66.67\%.

In~\cite{Dhall2015b}, Dhall et al. proposed to combine Action unit based face representation, low-level features for face and scene information for affective information on GAFF database. In face-level, they obtained the accuracy of 50.95\% using low-level features, while our method (INFA) obtained 58.33\%. It is shown that INFA performs better than their face-level approach on GAFF database. When they combined their features including Action unit based face representation, their system achieved the accuracy of 67.64\%, which is a little better than our multi-modal framework. It may be caused by that we use the designed parameters on HAPPEI database. 

\section{Conclusions}
\label{sec:conclusion}

In this paper, a multi-modal method combining face, upperbody and scene has been presented to analyze affect state of a group of people in an image. Firstly, for robustness we exploit three interesting information containing face, upperbody and scene in an image. In these information, face/upperbody is viewed as the bottom-up component while scene as top-down component. For representing an image, information aggregation was proposed to encode multiple people's information for group-level image. A robust multi-modal framework fusing face, upperbody and scene is finally presented to infer the affective state of a group of people.

We have conducted experiments on two challenging group-level emotion recognition databases. We show that INFA considerably improves the performance for group-level emotion recognition. Additionally, we evaluate multi-modal framework on HAPPEI and GAFF databases, respectively. Intensive experiments demonstrate that our multi-modal framework results in predicting the perceived group mood more accurately. 



%
%
%

\bibliographystyle{IEEEtran}
\bibliography{IEEEabrv,egbib2}




\end{document}